\documentclass[lettersize,journal]{IEEEtran}
\usepackage{amsmath,amsfonts}
\usepackage{algorithmic}
\usepackage{algorithm}
\usepackage{array}
\usepackage[caption=false,font=normalsize,labelfont=sf,textfont=sf]{subfig}
\usepackage{textcomp}
\usepackage{stfloats}
\usepackage{url}
\usepackage{verbatim}
\usepackage{graphicx}
\usepackage{cite}
\usepackage{color}
\usepackage{multirow}
\usepackage{amsmath}
\usepackage{booktabs} 
\usepackage{color,xcolor}
 
\begin{document}

\title{Weakly-Supervised 3D Scene Graph Generation via Visual-Linguistic Assisted Pseudo-labeling}
\author{Xu Wang,~\IEEEmembership{Member,~IEEE,}
        Yifan~Li,
        Qiudan~Zhang,~\IEEEmembership{Member,~IEEE,}
        Wenhui~Wu, ~\IEEEmembership{Member,~IEEE,}
        Mark Junjie Li, ~\IEEEmembership{Member,~IEEE,}
        Lin Ma ~\IEEEmembership{Senior Member,~IEEE,}
        Jianmin Jiang
\thanks{

Xu Wang, Yifan Li, Qiudan Zhang, Mark Junjie Li and Jianmin Jiang are with the College of Computer Science and Software Engineering, Shenzhen University, Shenzhen, 518060, China. Email: ( wangxu@szu.edu.cn, 2200271001@email.szu.edu.cn, qiudanzhang@szu.edu.cn, jj.li@szu.edu.cn and jianmin.jiang@szu.edu.cn).

Wenhui Wu is with the College of Electronics and Information Engineering, Shenzhen University, Shenzhen, 518060, China. Email: (wuwenhui@szu.edu.cn).

Lin Ma is with Meituan, Beijing, China. E-mail: forest.linma@gmail.com.
}
}






\markboth{Submitted to IEEE Transactions on Multimedia}%
{Shell \MakeLowercase{\textit{et al.}}: A Sample Article Using IEEEtran.cls for IEEE Journals}


\maketitle

\begin{abstract}
Learning to build 3D scene graphs is essential for real-world perception in a structured and rich fashion. However, previous 3D scene graph generation methods utilize a fully supervised learning manner and require a large amount of entity-level annotation data of objects and relations, which is extremely resource-consuming and tedious to obtain. To tackle this problem, we propose \textbf{3D-VLAP}, a weakly-supervised \textbf{3D} scene graph generation method via \textbf{V}isual-\textbf{L}inguistic \textbf{A}ssisted \textbf{P}seudo-labeling. Specifically, our 3D-VLAP exploits the superior ability of current large-scale visual-linguistic models to align the semantics between texts and 2D images, as well as the naturally existing correspondences between 2D images and 3D point clouds, and thus implicitly constructs correspondences between texts and 3D point clouds. First, we establish the positional correspondence from 3D point clouds to 2D images via camera intrinsic and extrinsic parameters, thereby achieving alignment of 3D point clouds and 2D images. Subsequently, a large-scale cross-modal visual-linguistic model is employed to indirectly align 3D instances with the textual category labels of objects by matching 2D images with object category labels. The pseudo labels for objects and relations are then produced for 3D-VLAP model training by calculating the similarity between visual embeddings and textual category embeddings of objects and relations encoded by the visual-linguistic model, respectively. Ultimately, we design an edge self-attention based graph neural network to generate scene graphs of 3D point cloud scenes. Extensive experiments demonstrate that our 3D-VLAP achieves comparable results with current advanced fully supervised methods, meanwhile significantly alleviating the pressure of data annotation.

\end{abstract}
\begin{IEEEkeywords}
3D scene graph generation, weakly-supervised learning
\end{IEEEkeywords}

\section{Introduction}
\IEEEPARstart{3}{D} scene graph generation has emerged as a pivotal topic in comprehensive scene understanding by constructing a directed graph-based abstract representation of objects and their relations in a scene. It offers a structured, enriched, and more intricate representation of 3D scenes compared to the traditionally used 3D point clouds or voxel grids, which is beneficial for a variety of downstream tasks, such as visual question answering~\cite{NuthalapatiVQA, azuma2022scanqa, etesam20223dvqa}, navigation~\cite{cai20223djcg, chen2020scanrefer, he2021transrefer3d,huang2023multi},  robotics~\cite{amiri2022reasoning,gadre2022continuous} and video compression~\cite{jiang2023video,zheng2024fuvc}. 


\begin{figure}[t]
 \begin{center}
  \includegraphics[width=0.4\textwidth]{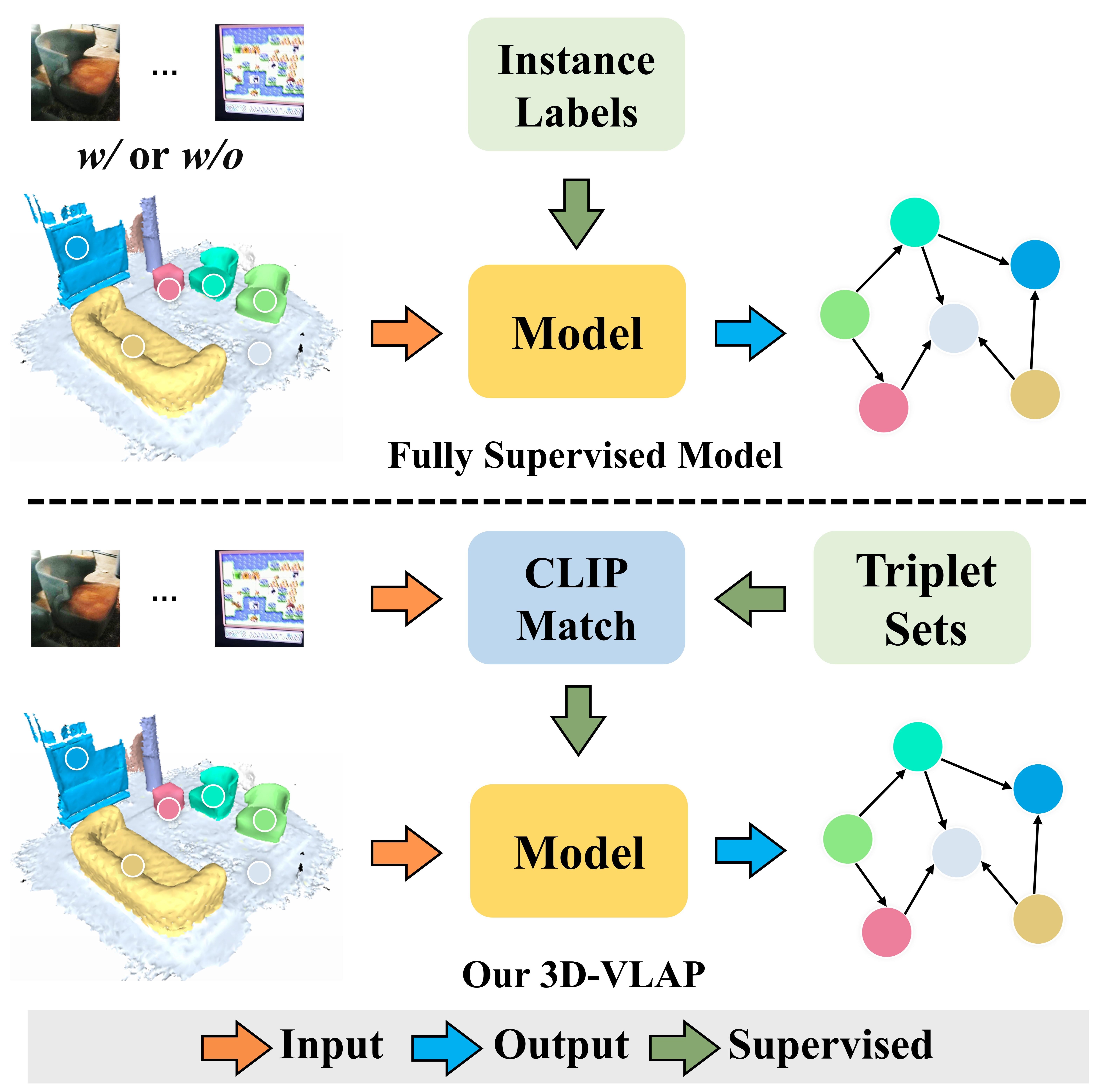}
  \end{center}
  \caption{\textbf{The comparison of fully supervised models and our 3D-VLAP.} Our designed 3D-VLAP utilizes a visual-linguistic model to assist in generating pseudo-labels for nodes and edges, thereby supervising the generation of 3D scene graphs. In contrast, fully supervised models directly predict 3D scene graphs supervised by \textit{instance-level} labels and may incorporate 2D images during training.}
  \label{figure_ws_framework}
\end{figure}

Substantial efforts have been invested in 3D scene graph generation~\cite{wu20223d, wald2020learning, wald2019rio, zhang2021knowledge, wang2023vl, lv2023revisiting, vaswani2017attention} within a fully supervised framework. However, these approaches necessitate extensive \textit{instance-level} label annotations for both objects and relations, which is label-intensive and cost-expensive. Consequently, scaling these methods to accommodate more intricate scenarios, object semantics, and relational predicates poses significant challenges. Naturally, the popular weakly supervised manner is exploited to generate scene graphs. Language~\cite{zhong2021learning, li2022integrating, ye2021linguistic, benetatos2023generating} and ungrounded scene graphs~\cite{shi2021simple, zareian2020weakly} are the mainstream signal forms for 2D weakly supervised methods. However, this weakly-supervised manner is not directly applicable to complex 3D scenes with a multitude of objects and relationships. In addition, two specific challenges are encountered in weakly-supervised 3D scene graph generation. Firstly, the 3D point cloud only provides coarse-grained structural information about objects, thereby complicating the inference of fine-grained semantic relationships among them. Secondly, in 3D scenes, the existence of numerous objects and their interrelations complicates the process of aligning category labels with visual information.

\begin{figure*}
\begin{center}
  \includegraphics[width=0.93\textwidth]{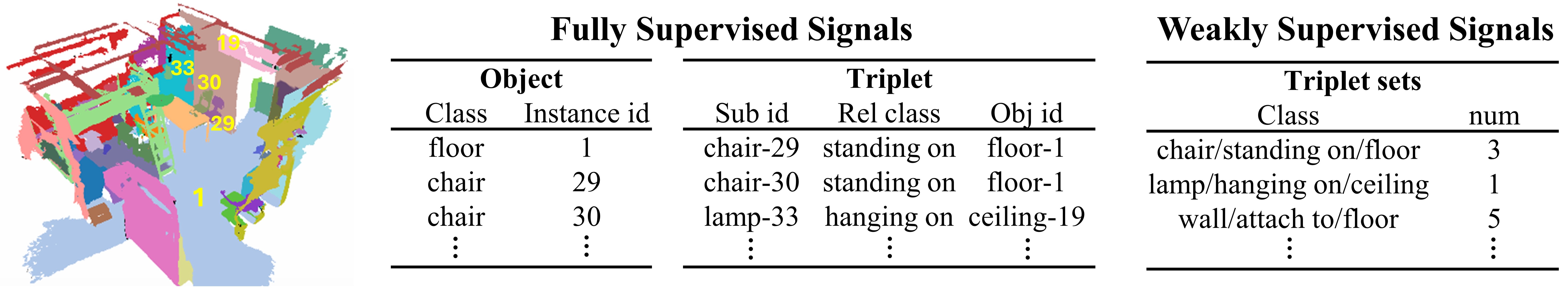}
  \end{center}
  \caption{\textbf{Comparison of fully supervised and our constructed weakly-supervised signals (\textit{triplet-set}).} In fully supervised annotation, each of the $K$ objects and $K\times(K-1)$ relations in the scene must be labeled individually. In contrast, our weakly-supervised annotation simplifies this process by identifying the triplets present in the scene and their count, thus significantly reducing the annotation cost.}
  \label{figure_triplet_set}
\end{figure*}

To tackle the above issues, we specify 2D images as a bridge connecting 3D point clouds and text category labels to construct weakly supervised signals for 3D scene graph generation. Hence, we develop 3D-VLAP, a weakly-supervised 3D scene graph generation method via visual-linguistic assisted pseudo-labeling. A comparison of our 3D-VLAP in a fully supervised manner is illustrated in Fig.~\ref{figure_ws_framework}. Concretely, we first establish the positional correspondences from 3D point clouds to 2D images via camera intrinsic and extrinsic parameters, thereby achieving alignment of 3D point clouds and 2D images. Subsequently, large-scale cross-modal visual linguistic model \textit{e.g.} CLIP is employed to aggregate visual embeddings of 2D images and text embeddings of object category labels, thereby indirectly aligning 3D point clouds with object category labels. Pseudo labels for objects and relations are generated by evaluating the similarity between their visual and text embeddings. A comparative analysis of fully supervised signals and our constructed weakly-supervised signals is illustrated in Fig.~\ref{figure_triplet_set}. However, directly deriving pseudo-labels for objects and relations through visual-linguistic interaction may result in imprecise relationship definitions, particularly in scenes with distinct instances of the same category. Thus, we propose a Hybrid Matching Strategy to improve the matching of text and visual embeddings of objects. Additionally, a Mask filter module is introduced to refine the generation of relation pseudo-labels used for model training. It is noteworthy that, since our pseudo-labels are dynamically generated, the quality of the pseudo-labels will improve as the model is optimized. Ultimately, the scene graph of the 3D scene is produced through reasoning by our designed edge self-attention based graph neural network (ESA-GNN). In summary, the main contributions of this paper are as follows:
\begin{itemize}
    \item We exploit the inherent characteristics of 3D scenes and the capabilities of the visual-linguistic model to construct weakly-supervised signals (\textit{triplet-set}) for 3D scene graph generation, thereby significantly alleviating the requirement for extensive human annotation.
    
    \item We introduce the inaugural weakly supervised methodology for 3D scene graph generation, termed 3D-VLAP, which capitalizes on large-scale visual-language models to facilitate the indirect alignment of 3D point clouds with object category labels. This approach empowers the generation of pseudo-labels for both objects and their relations.
    
    \item We conduct an in-depth analysis to illustrate the effectiveness of our designed 3D-VLAP model, which provides insightful information for future research on 3D weakly-supervised scene graph generation.
    
\end{itemize}

\section{Related Works}
\subsection{2D Scene Graph Generation}
Recently, an extensive body of literature has emerged in the field of 2D scene graph generation, with a significant emphasis placed on optimizing relational representations~\cite{zhang2019quaternion, liu2020hallucinating,9829320,zhang2023end}. For instance, Wang \textit {et al.}~\cite{10024789} moved beyond the conventional real-valued representations and utilized Quaternion Relation Embedding (QuatRE) to produce scene graphs with more expressive hyper-complex representations. To improve the tail predicates, Han \textit {et al.}~\cite{9829320} developed a Dual-Biased Predicate Predictor (DBiased-P) to boost the unbiased scene graph generation, which consists of a re-weighted primary classifier and an unweighted auxiliary classifier. Furthermore, Li \textit{et al.}~\cite{9726876} addressed the challenge of predicting unseen $predicates$ within triplets (\textit{i.e.}, $\langle subject$-$predicate$-$object \rangle$), proposing a novel zero-shot predicate prediction task that is crucial for enabling existing scene graph generation models to recognize novel relational categories. Nonetheless, these methods predominantly rely on fully supervised learning, necessitating densely annotated datasets that are both labor-intensive and costly to compile.
 

For the sake of alleviating the labeling efforts, several weakly supervised methods have been developed~\cite{zhang2017ppr, peyre2017weakly,li2022integrating}. Earlier research~\cite{zhang2017ppr, peyre2017weakly} predominantly relied on weakly supervised or pre-trained fully supervised object detectors~\cite{cinbis2016weakly, wang2014weakly, song2014learning, jie2017deep} to identify a set of RoIs (Region of Interest). The relationships between RoIs were determined by computing the selection scores and classification scores among the detected boxes. Yet, these methods have shown limited effectiveness in 2D scenes with multiple relations. However, these methods depend on off-the-shelf region proposal networks to localize objects in a scene, which are typically pre-trained on predefined fixed object detection datasets, thereby limiting the generalizability of locating unforeseen objects. Recently, with the emergence of cross-modal learning, it has become increasingly popular to utilize a visual-linguistic model to assist 2D scene graph generation with diverse relations~\cite{shi2021simple, zhong2021learning, ye2021linguistic}. They typically use pre-trained encoders to extract features from category labels and images, and then predict scene graphs based on the similarities between textual and visual embeddings. Furthermore, several 2D scene graph generation methods adopt ungrounded scene graphs~\cite{ozsoy2023location, shi2021simple} or languages~\cite{zhong2021learning, li2022integrating} as supervisory signals to optimize the training of models. A recent work~\cite{li2022integrating} introduced multi-instance learning (MIL) to enforce the model's perception of interactions between entities. However, due to the distinctive geometric complexity and underlying characteristics of 3D scenes, these methods are difficult to straightforwardly applied in the context of 3D scene graph generation.


\subsection{3D Scene Graph Generation}
More recent works~\cite{wu20223d, wald2020learning, wald2019rio,feng2023exploring} are devoted to leveraging Graph Convolutional Networks (GCN) to infer scene graphs of 3D scenes in a fully supervised fashion. For instance, Wald \textit{et al.}~\cite{wald2020learning} constructed a multi-scene indoor scene graph dataset named 3DSSG based on 3RScan point cloud data~\cite{wald2019rio}, where GCN serves as the learning baseline for analyzing semantic information within the scene. On this basis, Zhang \textit{et al.}~\cite{zhang2021exploiting} and Wu \textit{et al.}~\cite{wu2021scenegraphfusion} optimized the feature extraction of the initial relation, and integrated attention mechanism in GCN to improve node-edge interaction. Except for GCN optimization, Zhang \textit{et al.}~\cite{zhang2021knowledge} further leveraged category label topology to mitigate errors caused by visual similarity and perceptual noise during scene interpretation. Recently, Wang \textit {et al.}~\cite {wang2023vl} and Wu \textit {et al.}~\cite {wu2023incremental} have incorporated 2D images with rich semantic information into scene graph generation and used visual-language interaction to assist model training. Lv \textit{et al.}~\cite{lv2023revisiting} substituted the GCN model with Transformers~\cite{vaswani2017attention} for direct interaction among non-adjacent nodes and edges. However, they all rely on extensive instance-level annotations, which are extremely labor-intensive. Therefore, it has become particularly urgent to design a weakly supervised 3D scene generation method that significantly reduces annotation costs while maintaining prediction accuracy.

\section{The Proposed Method}
\begin{figure*}
  \centering 
  \includegraphics[width=\linewidth]{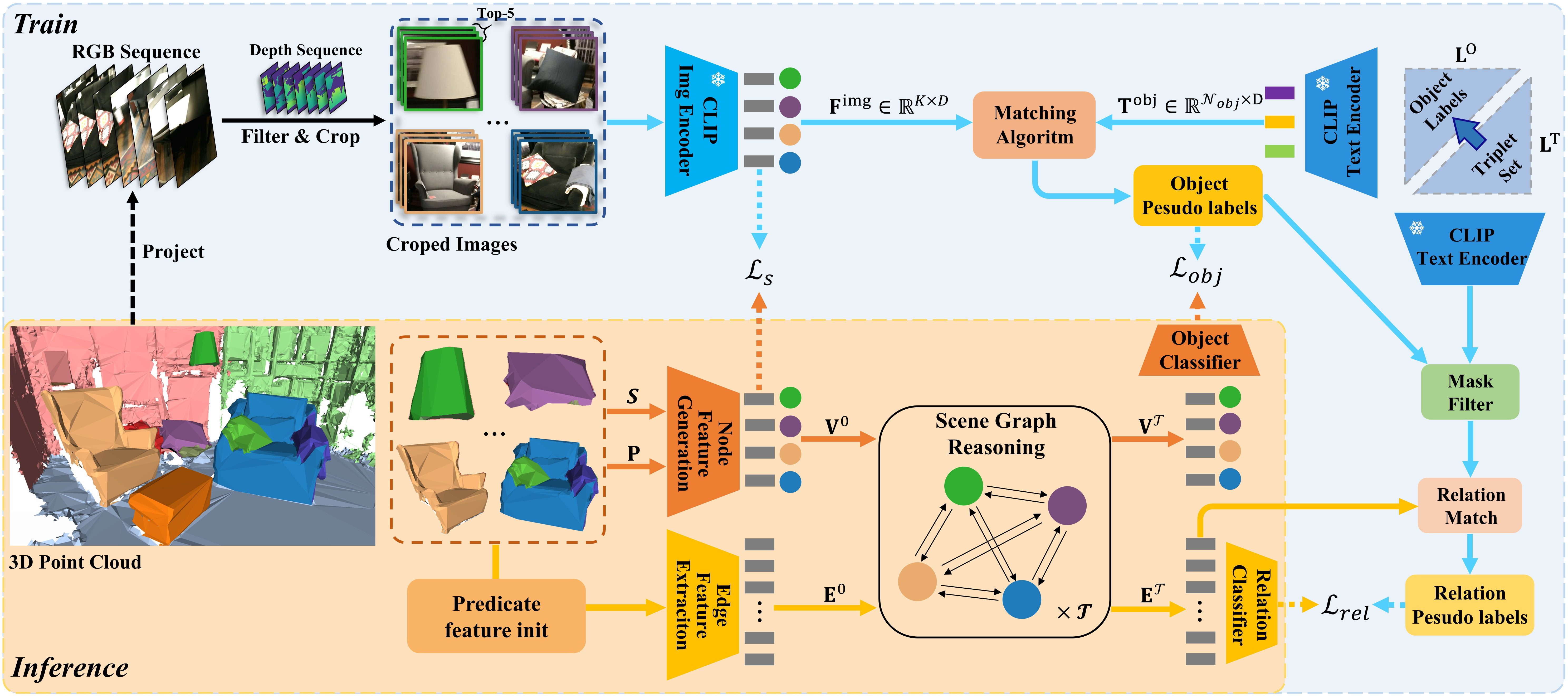}
  \caption{\textbf{A holistic architecture of semantic-enhanced weakly-supervised 3D scene graph generation via visual-linguistic driven pseudo-labeling.} We first utilize a Hybrid Matching Strategy to obtain pseudo-labels for each Instance during the training procedure. Then, when generating relational pseudo-labels, these pseudo-labels of objects are used to filter object pairs through a Mask Filter,  which facilitates the accuracy of relational pseudo-labels. Finally, these obtained pseudo-labels for objects and relations are assigned as supervisory signals during 3D scene graph model training. In the inference stage, we only need to feed 3D point cloud data into the proposed method to directly generate a 3D scene graph.}
  \label{fig:figure_3}
\end{figure*}

As illustrated in Fig.~\ref{fig:figure_3}, we develop a weakly supervised 3D scene graph generation approach with the help of pseudo-labels driven by visual-linguistic interactions. Specifically, we first exploit the intrinsic and extrinsic parameters of the camera to establish positional correspondences from 3D point clouds to 2D images. Subsequently, a large-scale visual-linguistic model is employed to extract the features of 2D images and category labels in text format. The 3D instances are then indirectly aligned to the category labels of the objects by matching the 2D images with the object category labels. Moreover, we also design a hybrid matching strategy to facilitate the matching of textual features and visual features of objects. Finally, the aligned visual and linguistic semantic representations are employed to assist the learning of 3D scene graph generation through pseudo-labeling, supervised by \textit{triplet-set}.


\subsection{Problem Formulation}
\label{problem_formulation}



Given a 3D point cloud $\mathbf{P}\in\mathbb{R}^{N \times 3}$ consisting of $N$ points, we assume there is a set of \textit{class-agnostic} instance masks $\mathbf{M}=\{\mathrm{M}_1,...,\mathrm{M_K}\}$ corresponding to $\mathrm{K}$ entities in $\mathbf{P}$. Our weakly supervised method for 3D scene graph generation refers to mapping the input 3D point cloud to a reliable semantically structured scene graph $\mathcal{G=\{O, R\}}$ under limited supervision signals~\cite{wald2020learning}. Particularly, we assign a triplet ( \textit{i.e.}, $\langle subject, predicate, object\rangle$ ) as the basic unit of a relationship. In graph $\mathcal{G}$, nodes $\mathcal{O} = \{o_i\}_{i=1...\mathrm{K}}$ denote 3D instances with category labels. Meanwhile, edges $\mathcal{R} = \{r_{ij}\}_{i\neq j,i,j=1...\mathrm{K}}$ describes the directional semantic and geometrical relations among triplet $\langle subject, predicate, object\rangle$, where $r_{ij}$ corresponds to the $predicate$, with the head node $o_i$ and tail node $o_j$ of $r_{ij}$ respectively corresponding to the $subject$ and $object$. 
Moreover, we denote the number of object categories in the dataset by $\mathcal{C}_{obj}$, the number of relationship categories by $\mathcal{C}_{rel}$, and the number of triplet categories by $\mathcal{C}_{tri}$, where $\mathcal{C}_{tri} =  \mathcal{C}_{obj} \times \mathcal{C}_{rel} \times \mathcal{C}_{obj}$. During model training, the only supervisory signal \textit{triplet-set} is termed as $\mathbf{L}^\mathrm{T} =\{l^t_i:n^t_i\}_{i=1...\mathcal{C}_{tri}}$. Each $l^t_i$ corresponds to a specific category within $\mathcal{C}_{tri}$, and $n^t_i$ signifies the frequency of occurrences of this category in the scene.

\subsection{3D Scene Graph Prediction Module}
\label{3D_sgg}

Herein, our devised 3D scene graph generation approach primarily comprises feature extraction of nodes and edges, as well as the scene graph reasoning module.

\smallskip
\noindent\textbf{Node Feature Extraction.}
As depicted in Fig.~\ref{fig:figure_3}, given a 3D point cloud $\mathbf{P}$ with instance masks $\mathbf{M}$, we first utilize the pre-trained PointNet~\cite{qi2017pointnet} to extract features of each point subset $\mathbf{P}_\mathrm{i}\in \mathbf{P}$, serving as the initial node embeddings. Each point subset $\mathbf{P}_\mathrm{i}$ corresponds to a distinct instance within the point cloud. To improve the model's perception of the position and direction of objects, we employ an MLP layer to produce the spatial properties $S=\{s_1, ..., s_\mathrm{K}\}$ for the bounding box $\mathrm{B} = \{b_1,...,b_\mathrm{K}\}$ of instances. Finally, the initial node embeddings $\mathbf{v}^0_i  \in \mathbb{R}^\mathrm{D}$ for the node $o_i$ are obtained by performing an addition operation on the point subset and spatial properties $S$. The formula is as follows:                                                                            
\begin{equation}
    \mathbf{v}^0_i = \mathtt{PointNet}(\mathbf{P}_\mathrm{i}) + \mathtt{MLP}(S),
    \label{eq:node_encoder}
\end{equation}
where $s_k = [b_x, b_y, b_z, \mu, \sigma, l, v]$ represents the spatial properties, and $b_k = \{b_x, b_y, b_z\}$ denotes the length, width, and height of the bounding box of the k-\textit{th} entity. $\mu$ and $\sigma$ represent the mean and standard deviation of the points in the subset $\mathbf{P}_\mathrm{i}$. $l$ indicates the longest side and $v = b_xb_yb_z$ denotes the volume of the bounding box. Finally, we obtain the node embeddings  $\mathbf{V}^0 = \{\mathbf{v}^0_i \in \mathbb{R}^\mathrm{D}\}_{i=1...\mathrm{K}}$ for the graph $\mathcal{G}$, where $D$ is the feature dimension.
\smallskip

\noindent\textbf{Edge Feature Extraction.}
To explore the relations among instances, we first construct the initial feature of edge $r_{ij}^{init}$ by calculating the representation difference between the bounding boxes of head node $o_i$ and tail node $o_j$. Subsequently, an MLP layer is adopted to update the initial spatial characteristics, and then produce the edge feature $e^0_{ij}$ for nodes $o_i$ and $o_j$. The formula is as follows,
\begin{equation}
    e^0_{ij} = \mathtt{MLP}(r^{init}_{ij}),
    \label{eq:2}
\end{equation}
where $r_{ij}^{init} = [\mu_i - \mu_j, \sigma_i - \sigma_j, b_i - b_j, ln\frac{l_i}{l_j}, \frac{v_i}{v_j}]$. $i$ and $j$ correspond to the subject $o_i$ and object $o_j$ in the triplet, respectively. Therefore, the initial edge embeddings $\mathbf{E}^0 = \{e^0_{ij}\in \mathbb{R}^\mathrm{D}\}_{i\neq j,i,j=1...\mathrm{K}}$ for the graph $\mathcal{G}$ is obtained.
\smallskip

\noindent\textbf{Scene Graph Reasoning.}
Inspired by the study in~\cite{wu2021scenegraphfusion}, after obtaining the features of nodes and edges, we design an edge self-attention based graph neural network (ESA-GNN) with $\mathcal{T}$ message passing layers, which facilitates the propagation and refinement of node and edge features by assimilating information from their respective neighborhoods. Since the perceptual field of edges in traditional GNNs is usually confined to nearby objects, we first introduce an \textbf{Edge self-attention} module to calculate long-range dependencies among edges, which treats the entire set of edge features as queries, keys, and values in Transformer's multi-head self-attention module. The formula is as follows:
\begin{equation}
    \mathbf{E}=[\mathrm{softmax}\left(\frac{\hat{Q}\hat{K}^T}{\sqrt{d_k}}\right)\hat{V}]^h_{i=1},
    \label{eq:self-attention}
\end{equation}
\begin{equation}
    \hat{Q} = \hat{g}^h_q(\mathbf{E}),\hat{K} = \hat{g}^h_q(\mathbf{E}),\hat{V} = \hat{g}^h_q(\mathbf{E}),
    \label{eq:qkv}
\end{equation}
where $h$ denotes the number of heads, and $d_k$ indicates the dimension of the features for Query $\hat{Q}$, Key $\hat{K}$ and Value $\hat{V}$. Subsequently, the ESA-GNN module employs Feature-wise Attention Network (FAN) modules~\cite{wu2021scenegraphfusion} to learn the coherence between the edges and nodes. For instance, given a triplet $\langle o_i, r_{ij}, o_j\rangle$, FAN calculates the attention between the features $[v_i, e_{ij}]$ and $v_j$, and then updates the features of node $v_i$ via an MLP layer. The formula is as follows:
\begin{equation}
    \mathrm{MFAT}(\mathbf{Q,T})=[\mathrm{FAT}(q_i, t_i)]^{h}_{i=1}=[\mathrm{S_{max}}(g_a(q_i))\odot t_i]^h_{i=1},
    \label{eq:mfat}
\end{equation}
\begin{equation}
    \mathrm{FAN}(v_i,e_{ij}, v_j)=\mathrm{MFAT}\left([g_q(v_i), g_e(e_{ij})], g_r(v_j)\right),
    \label{eq:fan}
\end{equation}
\begin{equation}
    v^{n+1}_i = g_v\left([v^n_i, \underset{j\in \mathcal{N}(i)}{max}(\mathrm{FAN}(v^n_i,e^n_{ij}, v^n_j)) ] \right),
    \label{eq:obj_gcn}
\end{equation}
\begin{equation}
    e_{ij}^{n+1} = g_e\left([v^n_i,e^n_{ij}, v^n_j] \right ),
    \label{eq:rel_gcn}
\end{equation}
where FAT indicates the feature-wise attention and MFAT denotes multi-head feature-wise attention operation. $\mathrm{S_{max}}$ indicates the softmax activation operation. $g_*$ represents the MLP operation, $1 \leq n \leq \mathcal{T}$ denotes the $n$-\textit{th} message passing layer, and $\mathcal{N}(i)$ represents the set of neighbor nodes of node $i$ in the graph. After completing the $\mathcal{T}$\textit{-th} message passing of relations between nodes, we obtain the final node embeddings $\mathbf{V}^\mathcal{T} = \{v^\mathcal{T}_i\in \mathbb{R}^\mathrm{D}\}$ and the edge embeddings $\mathbf{E}^\mathcal{T} = \{e^\mathcal{T}_{ij}\in \mathbb{R}^\mathrm{D}\}$. Finally, the embeddings $\mathbf{V}^\mathcal{T}$ and $\mathbf{E}^\mathcal{T}$ are fed into an object classifier to produce the final categories of objects and relations in the 3D point cloud scene. This object classifier contains a fully connected layer initialized with CLIP object embedding~\cite{Liao2022gen} and a relational classifier with the same structure as the classifier in PointNet~\cite{qi2017pointnet}. As a result, our designed scene graph reasoning module can effectively extract scene-level information and improve the prediction accuracy of relations by optimizing the capability of nodes and edges.

\subsection{Weakly-Supervised Learning via Triplets Set}
\label{weakly_sgg}


To indirectly optimize our designed 3D-VLAP method, we intend to utilize 2D images as a bridge to construct implicit matching relationships between 3D point clouds and category labels to generate weakly supervised signals. Elaborately, we first establish the positional correspondences between 3D instances in the point cloud data and 2D images via camera intrinsic and extrinsic parameters. Afterward, the powerful visual-linguistic interaction capability of the CLIP model~\cite{radford2021learning} is leveraged to construct pseudo-labels for nodes and edges, which are assigned as weak supervisory signals during the training procedure of our 3D-VLAP model.


\smallskip

\noindent\textbf{Pseudo-Labels Construction for Nodes.}
When constructing pseudo-labels for nodes, we first extract the object category set $\mathbf{L}^\mathrm{O} = \{l^o_i\}_{i=1...\mathcal{N}_{obj}}$ of instances from the \textit{triplet-set} $\mathbf{L}^\mathrm{T}$, where $\mathcal{N}_{obj}$ represents the category number of objects. Then, we project the point set $\mathbf{P}_\mathrm{i}$ of each instance to a 2D image sequence $\mathbf{I}$ based on camera intrinsic $\mathtt{Int}_p$ and extrinsic $\mathtt{Ext}_p$ parameters. The RGB and depth images corresponding to the point set $\mathbf{P}_\mathrm{i}$ of each instance are then selected. The projection formula is as follows,
\begin{equation}
    \frac{1}{Z_p}
\begin{bmatrix} 
U_p \\
V_p \\
1 \\
\end{bmatrix}
=
\begin{bmatrix}\mathtt{Int}_p\end{bmatrix}
\begin{bmatrix}\mathtt{Ext}_p\end{bmatrix}
\begin{bmatrix}
X_w \\
Y_w \\
Z_w \\
1
\end{bmatrix},
    \label{eq:projection}
\end{equation}
where $(X_w, Y_w, Z_w)$ represents the world coordinates of ${\mathbf{P}_\mathrm{i}}$ and $(U_p, V_p, Z_p)$ denotes the horizontal, vertical, and depth projections of the point corresponding to the image. After projection, we use the depth information to determine the effective projection points and select the top-5 2D images that can perfectly represent each instance in the 3D point cloud. These five images are then cropped according to the location of the projection points, allowing us to extract more precise visual features for each instance.

Subsequently, we adopt the pre-trained CLIP model to encode the selected five RGB images and the category set $\mathbf{L}^\mathrm{O}$ to obtain Image visual embeddings $\mathbf{F}^\mathrm{\mathrm{img}}  = \{f^\mathrm{img}_i\}_{i=1...\mathrm{K}}$ of each instance (average pooling of five images) and text embeddings $\mathbf{T}^\mathrm{obj}\in \mathbb{R}^{\mathcal{N}_{obj}\times \mathrm{D}}$ of labels, respectively.  However, directly leveraging CLIP models is incapable of generating accurate pseudo-labels for objects and relations, especially due to the confusion caused by different instances of the same category. Thus, we design a \textbf{Hybrid Matching Strategy} ($\mathtt{HMS}$) that combines the Hungarian algorithm with a Direct matching method to optimize the matching procedure of textual category embeddings and visual embeddings of objects. Specifically, we first construct a similarity matrix $S_m= \frac{\mathbf{T}^\mathrm{obj} \cdot \mathbf{F}^\mathrm{img}}{||\mathbf{T}^\mathrm{obj}|| \cdot ||\mathbf{F}^\mathrm{img}||}\in \mathbb{R} ^ {\mathcal{N}_{obj} \times K}$ to match the images and category labels, which calculates the cosine similarity between the textual embeddings of a category label and the visual embeddings of the corresponding image. Afterward, we apply the Hungarian algorithm to achieve one-to-one matching of textual category embeddings and visual embeddings of instances. For those unmatched instances, we use Direct matching to generate pseudo labels for them.

\begin{figure}[t]
  \centering 
  \includegraphics[width=\linewidth]{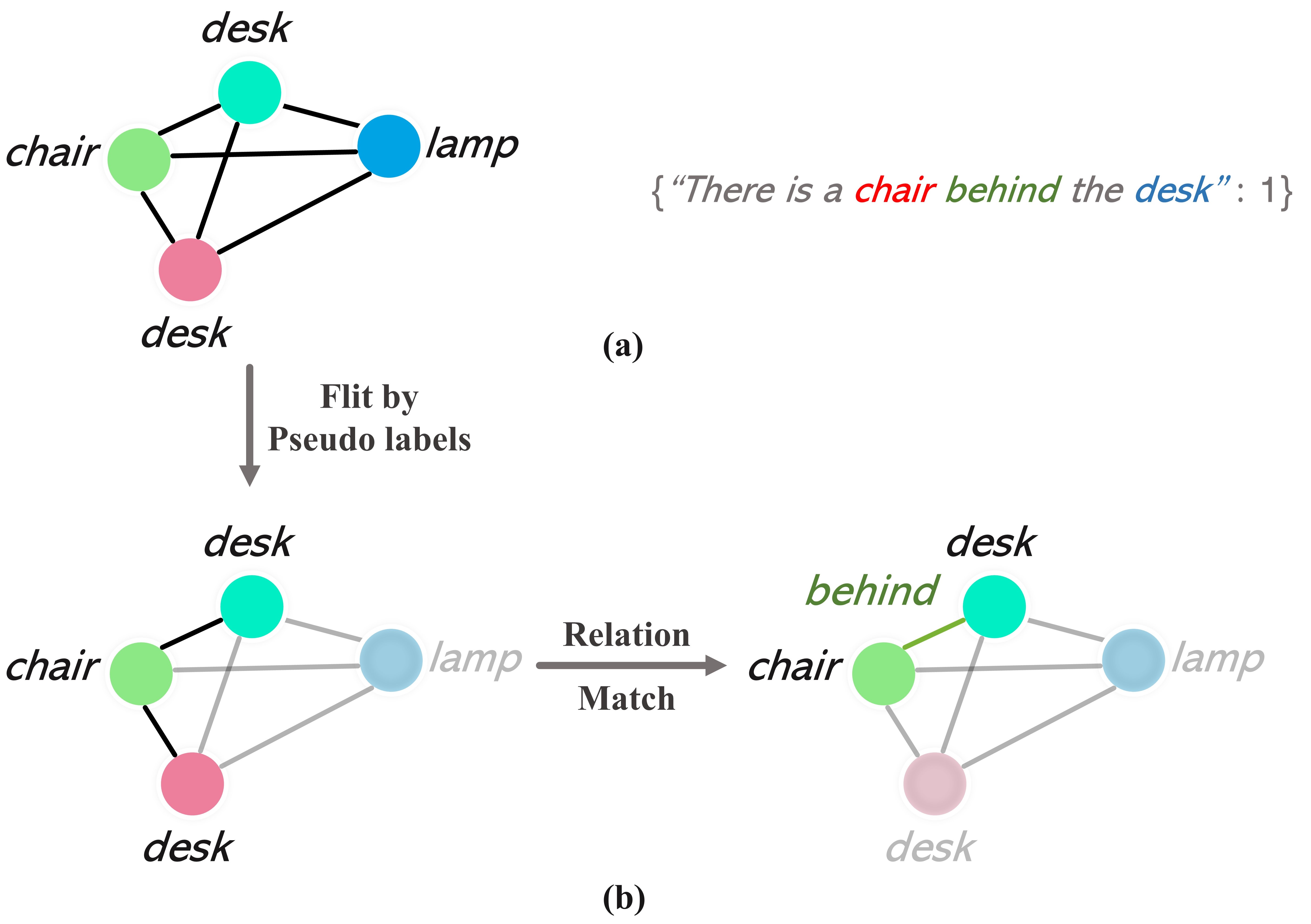}
  \caption{\textbf{Mask Filter.} The Mask Filter utilizes the pseudo-labels of nodes to filter the candidate edges in the scene graph, thereby optimizing the generation of relation pseudo-labels. (a) represents the scene graph with object pseudo-labels and the supervisory signal triplet set. In (b), the black edges represent the set of candidate edges, while the green edges represent the generated relation pseudo-labels.}
  \label{fig:maskfilter}
\end{figure}

\smallskip
\noindent\textbf{Pseudo-Labels Construction for Relations.}
Particularly, pseudo-label construction of predicates involves matching visual features of edges derived from point cloud data with corresponding triplet labels. As to each triplet $l^t_i$, we first prompt it according to the predicate type. For example, we use a template ``there is a [subject] on the [predicate] of [object]'' to represent the predicates like ``left'' and ``right''. For other predicates like ``attach to'' and ``close by'', the template is adjusted to ``there is a [subject] [predicate] [object]''. Later, the prompted results are fed into the text encoder of the CLIP to obtain textual embeddings $\mathbf{T}^\mathrm{tri}$ for each triplet. The pseudo-labels for edges are determined by calculating the similarity between textual embeddings $\mathbf{T}^\mathrm{tri}$ and edge embeddings $\mathbf{E}^\mathcal{T}$ filtered by the \textbf{Mask Filter} ($\mathtt{MF}$) module,
\begin{equation}
    \mathbf{Y}^\mathrm{R} = \mathtt{argmax}\left(  \frac{\mathbf{T}^\mathrm{tri} \cdot \mathtt{MF}(\mathbf{E}^\mathcal{T})}{||\mathbf{T}^\mathrm{tri}||\cdot ||\mathtt{MF}(\mathbf{E}^\mathcal{T})||} \right),
    \label{eq:rel_persudo_label}
\end{equation}
where $\mathbf{Y}^\mathrm{R}$ represents the pseudo-labels of edges. As illustrated in Fig.~\ref{fig:maskfilter}, the $\mathtt{MF}$ module selects the candidate edge sets based on the matching results of the subject and object in each triplet with the pseudo labels of the subject $o_i$ and object $o_j$ in the graph. It is worth noting that since the number of edges is significantly higher than the number of triplet labels, the pseudo labels of unmatched edges will be assigned the $None$.

Furthermore, to achieve precise alignment between the visual embeddings and textual embeddings for each triplet, we adopt a contrastive loss function to optimize the similarity calculation between the initial node embeddings $v^0$ and visual embeddings $f^\mathrm{img}$ of the image. The formula is as follows:
\begin{equation}
    \mathcal{L}_s=-\frac{1}{\mathrm{K}} \sum_{i\in \mathrm{K}} \left( \mathrm{log}\frac{\mathrm{exp}((v^0_i\cdot f^\mathrm{img}_i)/\tau)}{\sum_{j\in \mathrm{K}} \mathrm{exp}((v^0_i\cdot f^\mathrm{img}_i)/\tau)} \right),
    \label{eq:3}
\end{equation}
where $\tau$ represents the temperature hyper-parameter and is set to 0.1 in our experiment. By optimizing $\mathcal{L}_s$, we can facilitate feature alignment between point clouds and images, enabling subsequent acquisition of pseudo-labels for relations.

\begin{table*}[t]
\centering
\caption{Quantitative Evaluation of our proposed weakly supervised 3D scene graph generation method and the existing advanced fully supervised methods over datasets 3DSSG~\cite{wald2020learning} and 3DSSG-O27R16~\cite{zhang2021exploiting}.}
\label{table_compare_fully}
\resizebox{2\columnwidth}{!}{
\begin{tabular}{l|cc|ccc|cccccc|cccc}
\toprule \midrule
\multicolumn{1}{l}{} & \multirow{2}{*}{Supervision}   & \multirow{2}{*}{Model} & \multicolumn{3}{c|}{Object}  & \multicolumn{6}{c|}{Predicate}    & \multicolumn{4}{c}{Triplet}   \\ \cmidrule{4-16} 
 \multicolumn{1}{l}{}       &                      &                 & \multicolumn{1}{c|}{A@1}   & \multicolumn{1}{c|}{A@5}   & A@10  & \multicolumn{1}{c|}{A@1}   & \multicolumn{1}{c|}{A@3}   & \multicolumn{1}{c|}{A@5}   & \multicolumn{1}{c|}{mA@1}  & \multicolumn{1}{c|}{mA@3}  & mA@5  & \multicolumn{1}{c|}{A@50}  & \multicolumn{1}{c|}{A@100} & \multicolumn{1}{c|}{mA@50} & mA@100                     \\ \midrule
\multicolumn{1}{l}{\multirow{6}{*}[-2.2ex]{\rotatebox{90}{\normalsize 3DSSG}}} & \multirow{4}{*}{\begin{tabular}[c]{@{}c@{}}Fully\\ Supervised\end{tabular}} & SGPN~\cite{wald2020learning}                   & \multicolumn{1}{c|}{\underline{48.28}} & \multicolumn{1}{c|}{\underline{72.94}} & \underline{82.74} & \multicolumn{1}{c|}{91.32} & \multicolumn{1}{c|}{98.09} & \multicolumn{1}{c|}{99.15} & \multicolumn{1}{c|}{\underline{32.01}} & \multicolumn{1}{c|}{55.22} & 69.44 & \multicolumn{1}{c|}{87.55} & \multicolumn{1}{c|}{90.66} & \multicolumn{1}{c|}{\underline{41.52}} & \multicolumn{1}{c}{\underline{51.92}} \\
\multicolumn{1}{l}{}     &       & $\mathrm{SGG_{point}}$~\cite{zhang2021exploiting}            & \multicolumn{1}{c|}{51.42} & \multicolumn{1}{c|}{74.56} & 84.15 & \multicolumn{1}{c|}{92.40}  & \multicolumn{1}{c|}{97.78} & \multicolumn{1}{c|}{98.92} & \multicolumn{1}{c|}{\underline{27.95}} & \multicolumn{1}{c|}{49.98} & 63.15 & \multicolumn{1}{c|}{87.89} & \multicolumn{1}{c|}{90.16} & \multicolumn{1}{c|}{45.02} & \multicolumn{1}{c}{56.03} \\
\multicolumn{1}{l}{}    &        & SGFN~\cite{wu2021scenegraphfusion}                   & \multicolumn{1}{c|}{53.67} & \multicolumn{1}{c|}{77.18} & 85.14 & \multicolumn{1}{c|}{90.19} & \multicolumn{1}{c|}{98.17} & \multicolumn{1}{c|}{99.33} & \multicolumn{1}{c|}{41.89} & \multicolumn{1}{c|}{70.82} & 81.44 & \multicolumn{1}{c|}{89.02} & \multicolumn{1}{c|}{91.71} & \multicolumn{1}{c|}{58.37} & \multicolumn{1}{c}{67.61} \\
\multicolumn{1}{l}{}   &      & IMP~\cite{xu2017scene}                 & \multicolumn{1}{c|}{53.82} & \multicolumn{1}{c|}{77.34} & 86.09 & \multicolumn{1}{c|}{91.69} & \multicolumn{1}{c|}{98.33} & \multicolumn{1}{c|}{99.35} & \multicolumn{1}{c|}{46.71} & \multicolumn{1}{c|}{68.23} & 77.49 & \multicolumn{1}{c|}{89.26} & \multicolumn{1}{c|}{92.06} & \multicolumn{1}{c|}{55.51} & \multicolumn{1}{c}{64.62} 
                                                                            \\ 
\multicolumn{1}{l}{}  &     & VL-SAT~\cite{wang2023vl}                 & \multicolumn{1}{c|}{56.10} & \multicolumn{1}{c|}{77.80} & 85.90 & \multicolumn{1}{c|}{90.16} & \multicolumn{1}{c|}{98.47} & \multicolumn{1}{c|}{99.46} & \multicolumn{1}{c|}{50.55} & \multicolumn{1}{c|}{74.27} & 88.67 & \multicolumn{1}{c|}{90.29} & \multicolumn{1}{c|}{92.77} & \multicolumn{1}{c|}{64.73} & \multicolumn{1}{c}{74.12}  \\
                              \cmidrule(l){2-16} 
\multicolumn{1}{l}{} & \begin{tabular}[c]{@{}c@{}}Weakly\\ Supervised\end{tabular}            &   \textbf{3D-VLAP}           & \multicolumn{1}{c|}{\textbf{50.04}} & \multicolumn{1}{c|}{\textbf{74.02}} & 83.54 & \multicolumn{1}{c|}{87.52} & \multicolumn{1}{c|}{95.34} & \multicolumn{1}{c|}{97.99} & \multicolumn{1}{c|}{\textbf{35.19}} & \multicolumn{1}{c|}{48.39} & 62.32 & \multicolumn{1}{c|}{87.15} & \multicolumn{1}{c|}{90.04} & \multicolumn{1}{c|}{\textbf{42.70}} & \multicolumn{1}{c}{\textbf{52.82}} \\ \midrule \midrule

\multicolumn{1}{l}{\multirow{6}{*}[0ex]{\rotatebox{90}{\normalsize 3DSSG-O27R16}}} & \multirow{4}{*}{\begin{tabular}[c]{@{}c@{}}Fully\\ Supervised\end{tabular}} & SGPN~\cite{wald2020learning}                   & \multicolumn{1}{c|}{66.26} & \multicolumn{1}{c|}{92.16} & 98.04 & \multicolumn{1}{c|}{\underline{93.54}} & \multicolumn{1}{c|}{99.01} & \multicolumn{1}{c|}{99.57} & \multicolumn{1}{c|}{51.96} & \multicolumn{1}{c|}{71.19} & 83.67 & \multicolumn{1}{c|}{97.44} & \multicolumn{1}{c|}{98.32} & \multicolumn{1}{c|}{80.21} & 86.19  \\
\multicolumn{1}{l}{}     &        & $\mathrm{SGG_{point}}$~\cite{zhang2021exploiting} & \multicolumn{1}{c|}{66.39} & \multicolumn{1}{c|}{92.03} & \underline{97.22} & \multicolumn{1}{c|}{\underline{93.68}} & \multicolumn{1}{c|}{99.23} & \multicolumn{1}{c|}{\underline{99.04}} & \multicolumn{1}{c|}{47.98} & \multicolumn{1}{c|}{69.90} & 79.45 & \multicolumn{1}{c|}{97.65} & \multicolumn{1}{c|}{98.22} & \multicolumn{1}{c|}{80.89} & 86.49  \\
\multicolumn{1}{l}{}     &                  & SGFN~\cite{wu2021scenegraphfusion}                   & \multicolumn{1}{c|}{67.86} & \multicolumn{1}{c|}{93.16} & 98.54 & \multicolumn{1}{c|}{\underline{92.87}} & \multicolumn{1}{c|}{99.04} & \multicolumn{1}{c|}{99.60} & \multicolumn{1}{c|}{49.02} & \multicolumn{1}{c|}{73.67} & 83.85 & \multicolumn{1}{c|}{97.96} & \multicolumn{1}{c|}{98.78} & \multicolumn{1}{c|}{81.37} & 86.79  
                                                                            \\
\multicolumn{1}{l}{}     &                   & IMP~\cite{xu2017scene}                & \multicolumn{1}{c|}{67.96} & \multicolumn{1}{c|}{94.47} & 98.83 & \multicolumn{1}{c|}{\underline{93.87}} & \multicolumn{1}{c|}{99.03} & \multicolumn{1}{c|}{99.61} & \multicolumn{1}{c|}{48.29} & \multicolumn{1}{c|}{69.33} & 84.15 & \multicolumn{1}{c|}{97.19} & \multicolumn{1}{c|}{\underline{98.03}} & \multicolumn{1}{c|}{80.63} & \multicolumn{1}{c}{86.43} 
                                                                            \\
\multicolumn{1}{l}{}     &      & VL-SAT~\cite{wang2023vl}                  & \multicolumn{1}{c|}{67.25} & \multicolumn{1}{c|}{92.98} & 97.66 & \multicolumn{1}{c|}{\underline{90.99}} & \multicolumn{1}{c|}{98.09} & \multicolumn{1}{c|}{99.46} & \multicolumn{1}{c|}{61.84} & \multicolumn{1}{c|}{78.92} & 81.51 & \multicolumn{1}{c|}{97.25} & \multicolumn{1}{c|}{98.85} & \multicolumn{1}{c|}{87.95} & 89.62  \\ 
\cmidrule(l){2-16}
\multicolumn{1}{l}{}     &   \begin{tabular}[c]{@{}c@{}}Weakly\\ Supervised\end{tabular}              & \textbf{3D-VLAP}          & \multicolumn{1}{c|}{61.03} & \multicolumn{1}{c|}{91.09} & \textbf{97.28} & \multicolumn{1}{c|}{\textbf{93.90}} & \multicolumn{1}{c|}{98.73} & \multicolumn{1}{c|}{\textbf{99.50}} & \multicolumn{1}{c|}{47.58} & \multicolumn{1}{c|}{62.86} & 75.15 & \multicolumn{1}{c|}{96.91} & \multicolumn{1}{c|}{\textbf{98.04}} & \multicolumn{1}{c|}{64.20} & 72.58  \\ 
\midrule \bottomrule

\end{tabular}}
\end{table*}

\begin{table*}[t]
\centering
\caption{Comparison results of our proposed 3D-VLAP with eight advanced fully supervised approaches on scene graph classification (SGCls) and predicate classification (PredCls) tasks.}
\label{table_sgcls}
\resizebox{2\columnwidth}{!}{
\begin{tabular}{l|c|ccc|ccc}
\toprule \midrule
\multicolumn{1}{l}{} & \multirow{2}{*}{Model} & \multicolumn{3}{c|}{SGCls}  &  \multicolumn{3}{c}{PredCls}     \\ \cmidrule{3-8} 
\multicolumn{1}{l}{}       &    & \multicolumn{1}{c|}{R@20 / 50 / 100}    & \multicolumn{1}{c|}{ng-R@20 / 50 / 100} & \multicolumn{1}{c|}{mR@20 / 50 / 100} & \multicolumn{1}{c|}{R@20 / 50 / 100}    & \multicolumn{1}{c|}{ng-R@20 / 50 / 100} & mR@20 / 50 / 100
\\ \midrule
\multicolumn{1}{l}{\multirow{9}{*}[-2.0ex]{\rotatebox{90}{\normalsize 3DSSG}}} & Co-Occurrence~\cite{zhang2021knowledge}  & \multicolumn{1}{c|}{\underline{14.8} / \underline{19.7} / \underline{19.9}} & \multicolumn{1}{c|}{\underline{14.1} / \underline{20.2} / \underline{25.8}} & \multicolumn{1}{c|}{\underline{8.8} / \underline{12.7} / \underline{12.9}} & \multicolumn{1}{c|}{\underline{34.7} / \underline{47.4} / \underline{47.9}} & \multicolumn{1}{c|}{\underline{35.1} / \underline{55.6} / \underline{70.6}} & 33.8 / 47.4 / 47.9  \\
\multicolumn{1}{l}{} & KERN~\cite{chen2019knowledge}  & \multicolumn{1}{c|}{20.3 / 22.4 / 22.7} & \multicolumn{1}{c|}{20.8 / 24.7 / \underline{27.6}} & \multicolumn{1}{c|}{\underline{9.5} / \underline{11.5} / \underline{11.9}}  & \multicolumn{1}{c|}{\underline{46.8} / \underline{55.7} / \underline{56.5}} & \multicolumn{1}{c|}{\underline{48.3} / \underline{64.8} / \underline{77.2}} & \underline{18.8} / \underline{25.6} / \underline{26.5} \\
\multicolumn{1}{l}{} & SGPN~\cite{wald2020learning}  & \multicolumn{1}{c|}{27.0 / 28.8 / 29.0} & \multicolumn{1}{c|}{28.2 / 32.6 / 35.3} & \multicolumn{1}{c|}{19.7 / 22.6 / 23.1} & \multicolumn{1}{c|}{\underline{51.9} / \underline{58.0} / \underline{58.5}} & \multicolumn{1}{c|}{54.5 / 70.1 / 82.4} & 32.1 / \underline{38.4} / \underline{38.9}\\
\multicolumn{1}{l}{} & Schemata~\cite{sharifzadeh2021classification} & \multicolumn{1}{c|}{27.4 / 29.2 / 29.4} & \multicolumn{1}{c|}{28.8 / 33.5 / 36.3} & \multicolumn{1}{c|}{23.8 / 27.0 / 27.2} & \multicolumn{1}{c|}{\underline{48.7} / \underline{58.2} / \underline{59.1}} & \multicolumn{1}{c|}{\underline{49.6} / 67.1 / 80.2} & 35.2 / 42.6 / 43.3 \\
\multicolumn{1}{l}{} & Zhang \textit{et al.}~\cite{zhang2021knowledge} & \multicolumn{1}{c|}{28.5 / 30.0 / 30.1} & \multicolumn{1}{c|}{29.8 / 34.3 / 37.0} & \multicolumn{1}{c|}{24.4 / 28.6 / 28.8} & \multicolumn{1}{c|}{59.3 / 65.0 / \underline{65.3}} & \multicolumn{1}{c|}{62.2 / 78.4 / 88.3} & 56.6 / 63.5 / 63.8 \\
\multicolumn{1}{l}{} & IMP~\cite{xu2017scene}  & \multicolumn{1}{c|}{29.1 / 30.3 / 30.4} & \multicolumn{1}{c|}{31.6 / 39.4 / 44.7} & \multicolumn{1}{c|}{20.6 / 21.6 / 21.7} & \multicolumn{1}{c|}{69.4 / 79.3 / 80.0} & \multicolumn{1}{c|}{65.9 / 84.7 / 92.7} & 48.2 / 55.8 / 56.2 \\
\multicolumn{1}{l}{} & SGFN~\cite{wu2021scenegraphfusion}  & \multicolumn{1}{c|}{29.5 / 31.2 / 31.2} & \multicolumn{1}{c|}{31.9 / 39.3 / 45.0} & \multicolumn{1}{c|}{20.5 / 23.1 / 23.1} & \multicolumn{1}{c|}{65.9 / 78.8 / 79.6} & \multicolumn{1}{c|}{68.9 / 82.8 / 91.2} & 46.1 / 54.8 / 55.1 \\
\multicolumn{1}{l}{} & VL-SAT  ~\cite{wang2023vl}& \multicolumn{1}{c|}{32.0 / 33.5 / 33.7} & \multicolumn{1}{c|}{33.8 / 41.3 / 47.0} & \multicolumn{1}{c|}{31.0 / 32.6 / 32.7} & \multicolumn{1}{c|}{67.8 / 79.9 / 80.8} & \multicolumn{1}{c|}{70.5 / 85.0 / 92.5} & 57.8 / 64.2 / 64.3 \\ \cmidrule{2-8} 
\multicolumn{1}{l}{} & \textbf{3D-VLAP}              & \multicolumn{1}{c|}{\textbf{18.7} / \textbf{21.8} / \textbf{22.3}} & \multicolumn{1}{c|}{\textbf{16.8} / \textbf{22.9} / \textbf{27.8}} & \multicolumn{1}{c|}{\textbf{11.3} / \textbf{13.9} / \textbf{14.3}} & \multicolumn{1}{c|}{\textbf{53.5} / \textbf{64.4} / \textbf{65.7}} & \multicolumn{1}{c|}{\textbf{51.7} / \textbf{67.1} / \textbf{78.2}} & \textbf{31.8} / \textbf{41.2} / \textbf{41.9}  \\ \midrule \bottomrule
\end{tabular}
}

\end{table*}

\subsection{Loss Function}
To optimize the training procedure of our designed weakly-supervised 3D scene graph generation approach, the whole loss function $\mathcal{L}_{total}$ of the entire network is defined as follows,
\begin{equation}
    \mathcal{L}_{total} = \mathcal{L}_{obj} + \mathcal{L}_{rel} + \alpha * \mathcal{L}_{s},
    \label{eq:6}
\end{equation}
where $\mathcal{L}_{obj}$ and $\mathcal{L}_{rel}$ indicate the classification cross-entropy loss of nodes and edges between the predicted labels and the pseudo-labels respectively. $\mathcal{L}_s$ denotes the cosine similarity loss between the point cloud features of the object and the image embeddings. The $\alpha$ is set to 10.

\section{EXPERIMENTS}
\subsection{Experimental Setup}
\subsubsection{Datasets for Training and Testing}
We conduct the optimization and evaluation of the designed method over two 3D Scene Graph datasets: 3DSSG~\cite{wald2020learning} and 3DSSG-O27R16~\cite{zhang2021exploiting}. The 3D scenes for both datasets are derived from a large real-world indoor scene dataset 3RScan~\cite{wald2019rio}, and both datasets are appended with extensive relational labels. The 3DSSG includes 1553 3D structured indoor scenes with 160 object categories and 26 relation labels, while the 3DSSG-O27R16 consists of 27 object categories and 16 relation labels. In our experiment, we strictly adhere to the training and testing set partitioning, and data preprocessing procedures as described in datasets 3DSSG and 3DSSG-O27R16.

\subsubsection{Evaluation Metrics}
We employ the top-$k$ accuracy metric ($\mathrm{A@}k$) and mean top-$k$ accuracy metric ($\mathrm{mA@}k$)~\cite{wang2023vl} to estimate the prediction capability of our designed 3D-VLAP for objects, predicates and triplets. The higher the value of $\mathrm{A@}k$ and $\mathrm{mA@}k$, the stronger the model performance. Specifically, $\mathrm{A@}k$ measures the proportion of objects/predicates/triplets whose true label is ranked within the top-$k$ predictions, while $\mathrm{mA@}k$ calculates the average of $\mathrm{A@}k$ across all categories. In addition, we use the top-$k$ recall metric with graph constraints (R@$k$), the top-$k$ recall metric without graph constraints (ng-R@$k$), and the mean top-$k$ recall metric with graph constraints (mR@$k$)~\cite{zhang2021knowledge} for scene graph classification (SGCls) and predicate classification (PredCls). These evaluation metrics focus on ranking all predicates/triplets in a scene and then determining the proportion of correct ones among the top-$k$ ranked items.

\subsubsection{Implementation Details}
The proposed method is trained in an end-to-end manner. During training, we employ the Adam optimizer with cosine annealing learning rate decay as the optimizer. The learning rate and batch size are set to 0.0001 and 8, respectively. To extract features from images and text, we adopt the image and text encoders in the CLIP model~\cite{radford2021learning} in the experiment. In addition, the feature dimension $D$ is set to 512, and the number of the message passing layer $\mathcal{T}$ of the ESA-GNN is set to 2. Our proposed model is trained for 100 epochs on an NVIDIA GeForce A100 GPU.


\begin{figure*}[t]
  \centering 
  \includegraphics[width=\linewidth]{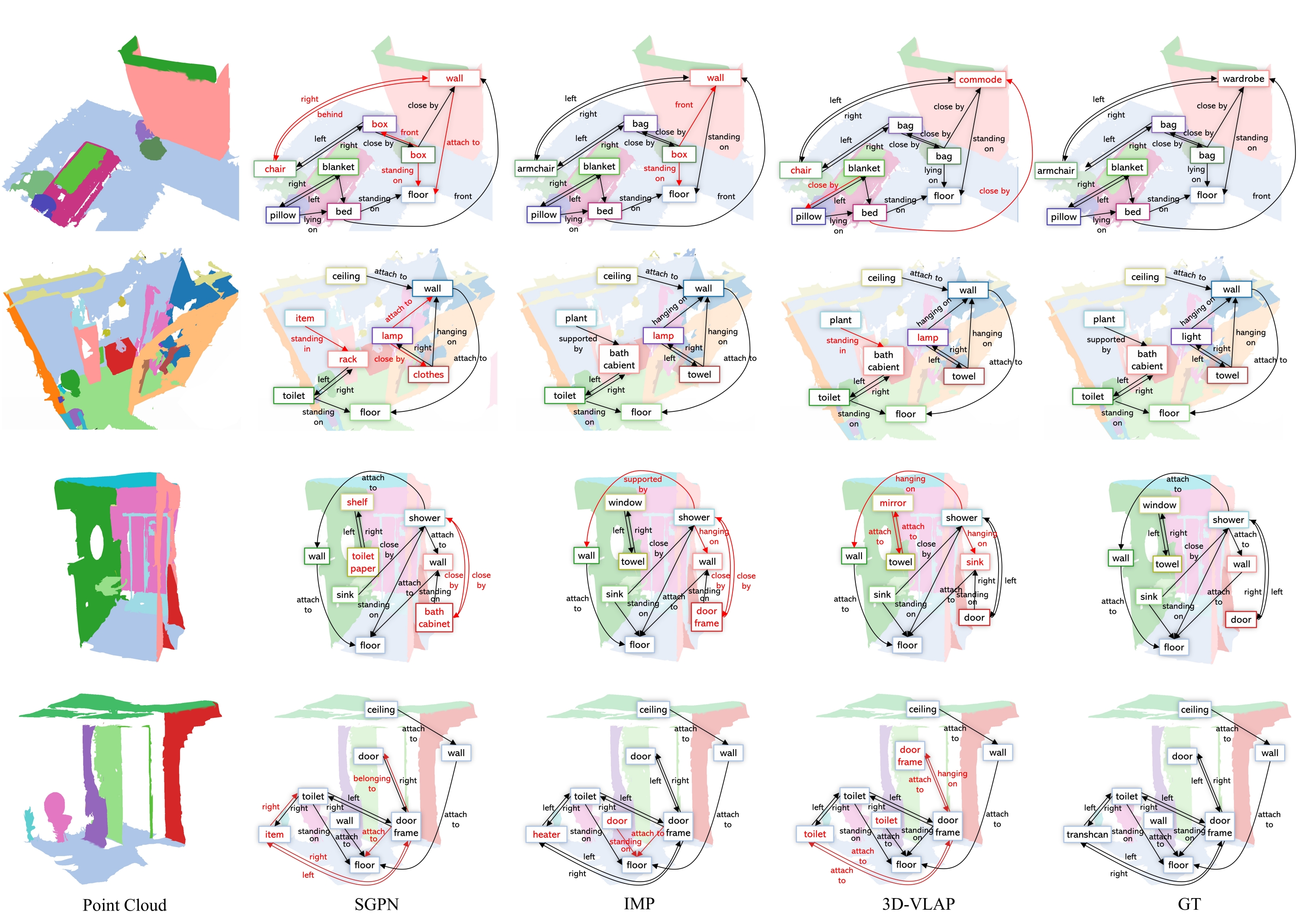}
  \caption{Qualitative comparison between our proposed method and two excellent fully supervised methods on the 3DSSG~\cite{wald2020learning} dataset, where red indicates misclassified nodes and edges.}
  \label{figure_qualitative_w_sgpn}
\end{figure*}

\subsection{Comparison with Fully-Supervised Methods}
To verify the performance of our designed weakly-supervised 3D scene graph generation method, we conduct comprehensive comparisons between our designed method and nine existing fully supervised 3D scene graph generation methods, including IMP~\cite{xu2017scene}, KERN~\cite{chen2019knowledge}, $\mathrm{SGG_{point}}$~\cite{zhang2021exploiting}, SGPN~\cite{wald2020learning}, Co-Occurrence~\cite{zhang2021knowledge}, SGFN~\cite{wu2021scenegraphfusion}, Schemata~\cite{sharifzadeh2021classification}, Zhang \textit{et al.}~\cite{zhang2021knowledge}, and VL-SAT~\cite{wang2023vl}.



In Table~\ref{table_compare_fully}, we provide the comparison results of our designed 3D-VLAP and five state-of-the-art fully supervised methods on the datasets 3DSSG~\cite{wald2020learning} and 3DSSG-O27R16~\cite{zhang2021exploiting}. From this table, we can observe that our designed weakly-supervised 3D-VLAP method is still inferior to the current excellent fully supervised models, but what's even more gratifying is that our method is already comparable to several fully supervised models in terms of specific indicators. For example, in triplet and object classification tasks, our method achieves comparable performance to SGPN~\cite{wald2020learning} on the 3DSSG dataset. This may be attributed to the integration of semantically rich image features in our weakly supervised signals, thereby enhancing the object recognition ability of the proposed model. Moreover, our method even outperforms the current SOTA model IMP~\cite{xu2017scene} in terms of the Predicate A@1 metric over the 3DSSG-O27R16 dataset. These comparison results illustrate that our proposed method obtains fine favorable performance for 3D scene graph generation in a weakly supervised fashion.

For the sake of illustrating the advantage of our designed method, we conduct two additional experiments on the 3DSSG dataset following the experimental settings described in~\cite{zhang2021knowledge}, including scene graph classification (SGCls) and predicate classification (PredCls). The comparison results of our proposed method and eight SOTA fully supervised methods are provided in Table~\ref{table_sgcls}. From Table~\ref{table_sgcls}, it can be clearly seen that our proposed method outperforms Co-Occurrence~\cite{zhang2021knowledge} by 3.9\% on metric R@20 and 2.7\% on metric ng-R@20, and achieves comparable results with KERN~\cite{chen2019knowledge} on the SGCls task. On the PredCls task, the performance of our proposed method exceeds methods SGPN~\cite{wald2020learning} and Schemata~\cite{sharifzadeh2021classification} by 1.6\% and 4.8\% on metric R@20, and exceeds 6.4\% and 6.2\% on metric R@50. Moreover, our proposed method achieves comparable results to SGPN~\cite{wald2020learning} and Schemata~\cite{sharifzadeh2021classification} on metrics ng-R@$k$ and mR@$k$. These findings demonstrate that our proposed method is capable of generating high-quality embeddings for relations even without using \textit{instance-level} annotations. 

\begin{figure*}
  \centering 
  \includegraphics[width=\linewidth]{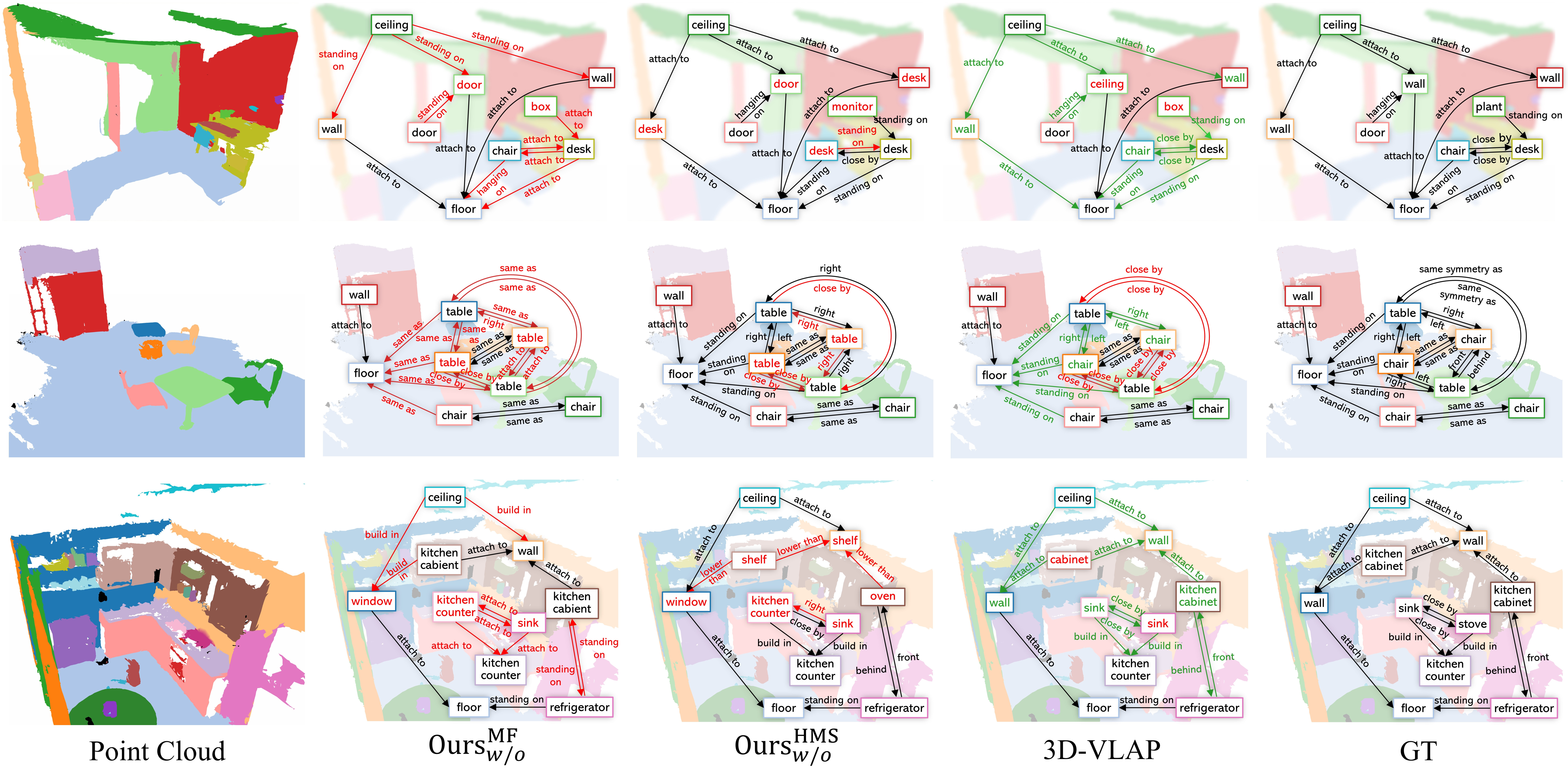}
  \caption{Qualitative results involved in ablation study on the 3DSSG~\cite{wald2020learning} dataset. We use black/red/green to represent edges and nodes that are correctly classified/misclassified/misclassified in method $\mathrm{Ours}_{w/o}^{\mathrm{MF}}$ or $\mathrm{Ours}_{w/o}^{\mathrm{HMS}}$ but correctly classified in our proposed full version method. $\mathrm{Ours}_{w/o}^{\mathrm{MF}}$ means to remove the Mask Filter, $\mathrm{Ours}_{w/o}^{\mathrm{HMS}}$ represents removing the Hybrid Matching Strategy, and GT indicates the Ground Truth of the scene graph.}
  \label{figure_qualitative_result}
\end{figure*}

Additionally, to further verify the effectiveness of our proposed weakly supervised 3D scene graph generation approach, we provide qualitative comparisons between our proposed method with two fully supervised approaches SGPN and IMP on the 3DSSG~\cite{wald2020learning} dataset in Fig.~\ref{figure_qualitative_w_sgpn}. As depicted in Fig.~\ref{figure_qualitative_w_sgpn}, we can see that the accuracy of scene graph produced by our designed approach is still inferior to other benchmark methods, such as the predicted predicate between the plant and bath cabinet in the second row and third column. This may be explained by the fact that the relational pseudo-labels generated by our method for these instances have low accuracy than the real labels of fully supervised methods. Moreover, we also observe that the category label ``commode'' by our method is closer to the ground truth ``wardrobe'' than the SGPN and IMP methods, as shown in the first row of the predicted scene graph results. This may be originated by the fact that our hybrid matching strategy better optimizes the matching procedure of textual category embeddings and visual embeddings of objects. These findings once again demonstrate that our proposed weakly supervised 3D scene graph generation method enables the production of pseudo label results for relations and instances that are comparable to fully supervised methods.

\subsection{Ablation Study}
\label{sec:Further_Exploration}
Herein, we conduct several ablation studies to investigate the effectiveness of different components in our 3D-VLAP approach over the 3DSSG~\cite{wald2020learning} dataset, and the comparison results are provided in Table~\ref{table_ablation_study}. By comparing the results from the fourth row to the last row in Table~\ref{table_ablation_study}, we discover that removing the classification cross-entropy loss of nodes $\mathcal{L}_{obj}$ has the greatest impact on the performance of our proposed method. In addition, comparing the results of the method removing the mask filter termed as $\mathtt{MF}$ and our proposed full-version method, we can find that removing the $\mathtt{MF}$ module leads to a decrease in prediction accuracy for predicates. This implies that our designed $\mathtt{MF}$ module can effectively utilize object pseudo-labels and exclude unmatched objects from triplet labels, thus improving the generation of pseudo-labels for predicates. Moreover, from Table~\ref{table_ablation_study}, we observe that removing the hybrid matching strategy termed as $\mathtt{HMS}$ causes a decline in prediction accuracy for objects and triplets. This indicates that the $\mathtt{HMS}$ module is playing a quite significant role in 3D scene graph generation in our proposed method. Comparing all the baseline methods with our designed full-version method, we can see that our whole method achieves superior performance. This again provides evidence that the complementary coherence among these distinct components can facilitate the performance of weakly supervised 3D scene graph generation.
\begin{table}[t]
\centering
\caption{Quantitative results of distinct components involved in ablation study on the 3DSSG~\cite{wald2020learning} dataset, where $\mathtt{HMS}$ indicates the hybrid matching strategy and $\mathtt{MF}$ denotes the mask filter.}
\label{table_ablation_study}
\resizebox{\columnwidth}{!}{
\begin{tabular}{ccccc|cc|cc|cc}
\toprule
\multirow{2}{*}{$\mathcal{L}_{obj}$} & \multirow{2}{*}{$\mathcal{L}_{rel}$} & \multirow{2}{*}{$\mathtt{HMS}$}& \multirow{2}{*}{$\mathtt{MF}$} & \multirow{2}{*}{$\mathcal{L}_{s}$} & \multicolumn{2}{c|}{Object} & \multicolumn{2}{c|}{Predicate} & \multicolumn{2}{c}{Triplet} \\ \cmidrule{6-11} 
                                     &                                      &                     &                                    &          & A@5          & A@10         & mA@3           & mA@5          & mA@50        & mA@100       \\ \midrule
$\surd$                              &                                      &                     &                                    &           & 67.12        & 78.00        & 9.61           & 26.98         & 28.38        & 40.07        \\
$\surd$                              & $\surd$                              &                     &                                    &          & 67.32        & 78.52         & 19.75          & 40.32         & 27.18        & 38.57       \\
$\surd$                              & $\surd$                              & $\surd$             &                                    &           & 72.61        & 82.09        & 21.26          & 40.78         & 33.31        & 43.76        \\
$\surd$                              & $\surd$                              & $\surd$             &$\surd$                            &           & 72.32        & 82.68        & 44.53          & 60.31         & 41.86        & 52.82        \\
\midrule
     &       $\surd$               &         $\surd$     &     $\surd$                  &     $\surd$       &    20.52   &   24.95        &   44.97     &   59.63    & 0.78   & 13.67        \\
$\surd$        &      &         $\surd$   &   $\surd$                    &   $\surd$  & 73.65        & 81.77         & 10.46          & 27.54         & 35.78        & 47.45       \\
$\surd$               & $\surd$     &          &     $\surd$     &    $\surd$    & 65.62        & 77.34        & 45.56          & 61.54         & 38.93        & 49.15        \\
$\surd$                              & $\surd$                              & $\surd$             &           &    $\surd$    & \textbf{74.81}        & \textbf{83.77}        &  21.08          & 40.12         & 31.98        & 44.33        \\
$\surd$                              & $\surd$                              & $\surd$             & $\surd$           &   $\surd$    & 74.02        & 83.54       & \textbf{48.39}          & \textbf{62.32}         & \textbf{42.70}        & \textbf{52.82}        \\
 \bottomrule
\end{tabular}
}
\end{table}

To better demonstrate the effectiveness of the distinct components in our designed 3D-VLAP, we carefully specify a series of experimental settings including the method that removes the mask filter termed as $\mathrm{Ours}_{w/o}^{\mathrm{MF}}$, the method that removes the hybrid matching strategy termed as $\mathrm{Ours}_{w/o}^{\mathrm{HMS}}$ and our proposed full-version method. The qualitative results are depicted in Fig.~\ref{figure_qualitative_result}. From this figure, it is obvious that the prediction accuracy of edges by the method $\mathrm{Ours}_{w/o}^{\mathrm{MF}}$ is significantly lower than the method $\mathrm{Ours}_{w/o}^{\mathrm{HMS}}$, and the accuracy of node classification results by the method $\mathrm{Ours}_{w/o}^{\mathrm{MF}}$ is still inferior to the method $\mathrm{Ours}_{w/o}^{\mathrm{HMS}}$. This may be caused by the fact that the hybrid matching strategy can effectively improve the accuracy of pseudo-label construction for instances, and the mask filter enables the enhancement of the accuracy of relational pseudo-label construction. Moreover, by comparing the scene graph results of the methods $\mathrm{Ours}_{w/o}^{\mathrm{MF}}$, $\mathrm{Ours}_{w/o}^{\mathrm{HMS}}$ and our proposed full version method, we find that our method achieves higher node and edge classification accuracy than using either module alone. This may be due to the fact that incorrect pseudo-labels of nodes will cause confusion in the construction of edge pseudo-labels. Similarly, incorrect edge pseudo-labels will also affect node pseudo-labels, such as affecting the overall performance of the method.

\begin{table*}[t]
\renewcommand\arraystretch{1.13}
\centering
\caption{Relation category grouping based on the number of samples in the training set.}
\label{table_split}
\resizebox{2\columnwidth}{!}{
\begin{tabular}{c|c}
\hline
Split & Predicates                                                                                                                                        \\ \hline
Head  & left, right, front, behind, close by, same as, attached to, standing on                                                                          \\
Body  & bigger than, smaller than, higher than, lower than, lying on, hanging on                                                                         \\
Tail  & supported by, inside, same symmetry as, connected to, leaning against, part of, belonging to, build in, standing in, cover, lying in, hanging in \\ \hline
\end{tabular}
}
\end{table*}
\begin{figure}[t]
  \centering 
  \includegraphics[width=\linewidth]{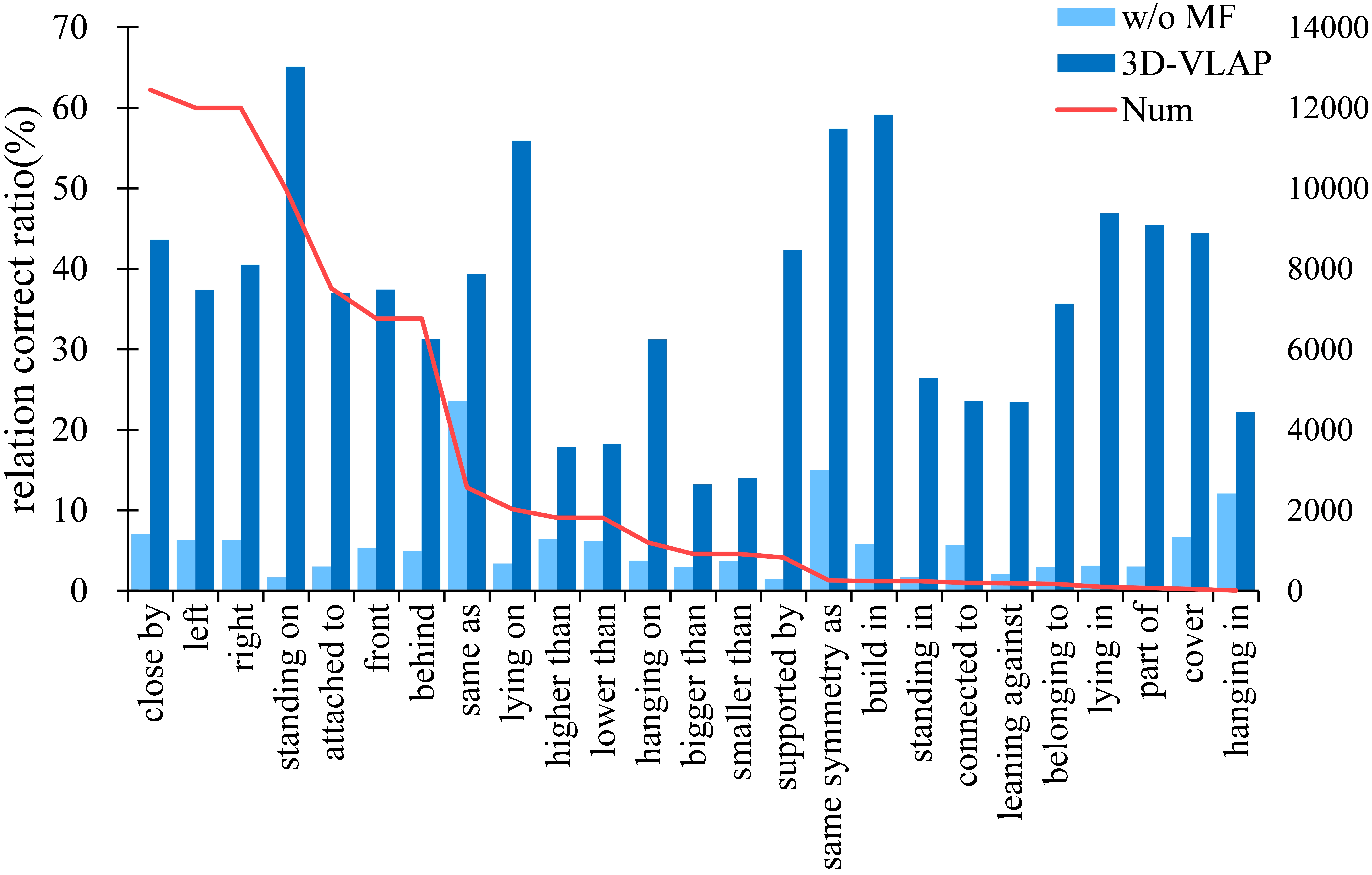}
  \caption{\textbf{Accuracy of relational pseudo-label generation.} The red line represents the number of each relationship type in the training set, while the w/o MF represents the removal of the mask filter module. Accuracy is calculated as the number of correctly classified pseudo-labels divided by the total number of true labels.}
  \label{figure_4}
\end{figure}
\subsection{More Analysis of Our 3D-VLAP Model}
\subsubsection{Quality of Predicate Pseudo Labels}
The quality of pseudo labels directly affects the performance of our 3D-VLAP model. To illustrate the quality of our constructed pseudo labels, we present the accuracy of the obtained pseudo labels across various relational categories in Fig.~\ref{figure_4}. From this figure, it can be clearly seen that the accuracy of the pseudo labels for relations is not affected by the number of predicates in the training set. It is worth noting that even with a limited number of samples, our proposed 3D-VLAP model can generate pseudo-labels with high accuracy for categories with rich semantic content. Moreover, we also observe the introduction of the mask filter module significantly improves the accuracy of the constructed relational pseudo-labels. This also confirms that the mask filter module contributes to a substantial increase in average category accuracy by 29.72\%, confirming its vital role in boosting the model's relationship prediction capabilities.


\subsubsection{Comparisons of Head, Body and Tail Predicates}
As shown in Table~\ref{table_split}, we first divide the 26 predicate categories into three groups: \textit{Head}, \textit{Body} and \textit{Tail} according to the number of predicates. Subsequently, to illustrate the effectiveness of 3D-VLAP in rare relation prediction, we compare the prediction accuracy of 3D-VLAP and SGFN on different predicate groups in terms of metrics mA@3 and mA@5. The comparison results are provided in Table~\ref{head_body_tail}. From this table, it is obvious that our model performs even better in the Tail group than the method SGPN. This indicates that our proposed model can effectively capture the semantic information of small-sample relation categories.

\begin{table}[t]
\centering
\caption{Performance comparison of three groups of predicate categories Head, Body and Tail predicted by our 3D-VLAP model and SCPN~\cite{wald2020learning} model respectively.}
\label{head_body_tail}
\resizebox{\columnwidth}{!}{
\begin{tabular}{c|cccccc}
\toprule
\multirow{3}{*}{Model} & \multicolumn{6}{c}{Predicate}                                                                                                                    \\ \cmidrule{2-7} 
                       & \multicolumn{2}{c|}{Head}                             & \multicolumn{2}{c|}{Body}                             & \multicolumn{2}{c}{Tail}         \\ \cmidrule{2-7} 
                       & \multicolumn{1}{c|}{mA@3} & \multicolumn{1}{c|}{mA@5} & \multicolumn{1}{c|}{mA@3} & \multicolumn{1}{c|}{mA@5} & \multicolumn{1}{c|}{mA@3} & mA@5 \\ \midrule
SGPN~\cite{wald2020learning}       & \multicolumn{1}{c|}{96.2} & \multicolumn{1}{c|}{99.0} & \multicolumn{1}{c|}{66.2} & \multicolumn{1}{c|}{86.7} & \multicolumn{1}{c|}{10.2} & 28.4 \\ \midrule
3D-VLAP         & \multicolumn{1}{c|}{83.3} & \multicolumn{1}{c|}{95.6} & \multicolumn{1}{c|}{31.9} & \multicolumn{1}{c|}{60.2} & \multicolumn{1}{c|}{\textbf{31.9}} & \textbf{39.1} \\ \bottomrule
\end{tabular}
}
\end{table}

\subsubsection{Portability Analysis of Our Weakly-Supervised Framework}
Our designed 3D-VLAP method is composed of two principle components: weakly-supervised pseudo-label generation (denoted as $\mathrm{3D}$-$\mathrm{VLAP^\dagger}$) and 3D scene graph generation. To evaluate the portability of our designed 3D-VLAP method, we substitute the training operations of traditional fully supervised 3D scene graph generation models (SGPN, IMP and SGFN) with our weakly supervised framework, thereby extending its application to the weakly supervised domain. The comparisons between the extended methods and our 3D-VLAP are provided in Table~\ref{table_ws_framework}. From Table~\ref{table_ws_framework}, we observe that the performance of the extended methods (SGPN+$\mathrm{3D}$-$\mathrm{VLAP^\dagger}$, IMP+$\mathrm{3D}$-$\mathrm{VLAP^\dagger}$ and SGFN+$\mathrm{3D}$-$\mathrm{VLAP^\dagger}$) is slightly inferior to their fully supervised version. Considering that weakly supervised learning methods can reduce the burden of label annotation, we believe that this trade-off between performance and cost is acceptable. Moreover, we find that our 3D-VLAP achieves superior performance than the extended methods. This also indirectly illustrates the excellence of our proposed weakly supervised 3D scene graph generation framework.
\begin{table}[t]
\centering
\caption{Performance comparison of different 3D scene graph generation methods in weakly and fully supervised versions.}
\label{table_ws_framework}
\resizebox{\columnwidth}{!}{
\begin{tabular}{c|cc|cc|cc}
\toprule
\multirow{2}{*}{Model} & \multicolumn{2}{c|}{Object}        & \multicolumn{2}{c|}{Predicate}     & \multicolumn{2}{c}{Triplet}         \\ \cmidrule{2-7} 
                       & \multicolumn{1}{c|}{A@1}   & A@5   & \multicolumn{1}{c|}{mA@1}  & mA@3  & \multicolumn{1}{c|}{mA@50} & mA@100 \\ \midrule
SGPN~\cite{wald2020learning}                   & \multicolumn{1}{c|}{48.28} & 72.94 & \multicolumn{1}{c|}{32.01} & 55.22 & \multicolumn{1}{c|}{41.52} & 51.92  \\
SGPN +$\mathrm{3D}$-$\mathrm{VLAP^\dagger}$                    & \multicolumn{1}{c|}{42.26} & 64.85 & \multicolumn{1}{c|}{25.40} & 45.08 & \multicolumn{1}{c|}{38.86} & 49.62  \\ \midrule
IMP~\cite{xu2017scene}                   & \multicolumn{1}{c|}{53.82} & 77.34 & \multicolumn{1}{c|}{46.71} & 68.23 & \multicolumn{1}{c|}{55.51} & 64.62  \\
IMP +$\mathrm{3D}$-$\mathrm{VLAP^\dagger}$                    & \multicolumn{1}{c|}{47.09} & 71.92 & \multicolumn{1}{c|}{30.00} & 46.50 & \multicolumn{1}{c|}{42.51} & 54.54  \\ \midrule
SGFN~\cite{wu2021scenegraphfusion}                   & \multicolumn{1}{c|}{53.67} & 77.18 & \multicolumn{1}{c|}{41.89} & 70.82 & \multicolumn{1}{c|}{58.37} & 67.61  \\
SGFN +$\mathrm{3D}$-$\mathrm{VLAP^\dagger}$               & \multicolumn{1}{c|}{48.83} & 74.12 & \multicolumn{1}{c|}{30.65} & 46.22 & \multicolumn{1}{c|}{40.81} & 50.08  \\  \midrule
3D-VLAP               & \multicolumn{1}{c|}{50.04} & 74.02 & \multicolumn{1}{c|}{35.19} & 48.39 & \multicolumn{1}{c|}{42.70} & 52.82  \\  \bottomrule
\end{tabular}
}
\end{table}
\subsubsection{Analysis of Different Supervisory Signals}
Our 3D-VLAP model exploits rich semantic information in 2D images to enhance the model's training procedure and improve the accuracy of 3D scene graph generation. To further verify the effectiveness of our constructed weakly-supervised signals for 3D scene graph generation, we carefully specify a series of experimental settings in terms of different supervised signals, including the method that uses \textit{instance-level} ground truth labels for nodes and edges termed as $\mathrm{3D}$-$\mathrm{VLAP}^{1}$, the method that uses our designed triplet supervisory signal and \textit{instance-level} ground truth labels for edges termed as $\mathrm{3D}$-$\mathrm{VLAP}^{2}$, the method that uses our designed triplet supervisory signal and \textit{instance-level} ground truth labels for nodes termed as $\mathrm{3D}$-$\mathrm{VLAP}^{3}$. It is worth noting that the method $\mathrm{3D}$-$\mathrm{VLAP}^{1}$ is a fully supervised method because both \textit{instance-level} ground truth labels for nodes and edges are used as its supervision signals. The comparison results of these methods using different supervisory signal and our proposed 3D-VLAP method are provided in Table~\ref{table_signals}. From this table, we can observe that the performance of $\mathrm{3D}$-$\mathrm{VLAP}^{1}$ model obtains superior performance than other methods. By comparing the method $\mathrm{3D}$-$\mathrm{VLAP}^{1}$ and fully supervised method SGFN, we can see that the performance of SGFN is inferior to $\mathrm{3D}$-$\mathrm{VLAP}^{1}$. This indicates that our proposed model can also achieve quite excellent performance when extended to the fully supervised version. In addition, comparing the method $\mathrm{3D}$-$\mathrm{VLAP}^{2}$ and our 3D-VLAP, we can find that our 3D-VLAP obtains better performance than $\mathrm{3D}$-$\mathrm{VLAP}^{2}$. This illustrates that our constructed triplet supervisory signal is capable of facilitating our model to generate accurate scene graphs for 3D scenes.


\section{CONCLUSION}
In this paper, we tackle the difficult task of annotating 3D scene graphs from a novel perspective. Specifically, we propose a first weakly-supervised method tailored for 3D scene graph generation, which exploits the visual-linguistic interaction capability of CLIP to implicitly construct correspondences between texts and 3D point clouds with no need for fine-grained annotations. To indirectly supervise the generation of 3D scene graphs, a hybrid matching strategy and a mask filtering algorithm are introduced to improve the quality of pseudo-label generation. Furthermore, our designed weakly supervised framework can be applied briefly and efficiently to convert fully supervised 3D scene graph models into weakly supervised versions. Extensive experimental results demonstrate that our 3D-VLAP obtains comparable results to most existing fully supervised SOTA methods, while significantly alleviating the burden of label annotations. We believe that our method provides valuable insights for future research on weakly-supervised 3D scene graph generation.
\begin{table}[t]
\centering
\caption{Performance comparison of 3D-VLAP under different supervisory signals. $\mathrm{L}^\mathrm{T}$ represents our devised \textit{triplet-set} supervisory signal, $\mathrm{GT}^\mathrm{O}$ and $\mathrm{GT}^\mathrm{P}$ denote the \textit{instance-level} ground truth labels for nodes and edges, respectively. RA indicates the best results of the random test.}
\label{table_signals}
\resizebox{\columnwidth}{!}{

\begin{tabular}{c|ccc|c|c|c}
\toprule
\multirow{2}{*}{Model}           & \multicolumn{3}{c|}{Signals} & Object         & Predicate      & Triplet   \\ 
\cmidrule{2-7}      &   $\mathrm{L}^\mathrm{T}$    &     $\mathrm{GT}^\mathrm{O}$   &   $\mathrm{GT}^\mathrm{P}$  & A@1/5/10   & mA@1/3/5       & mA@50/100 \\ \midrule
\begin{tabular}[c]{@{}c@{}}SGFN~\cite{wu2021scenegraphfusion} \\ (baseline)\end{tabular}    &                 & $\surd$     & $\surd$                   & 53.7/77.2/85.1 & 41.9/70.8/81.4 & 58.4/67.6 \\ \midrule
$\mathrm{3D}$-$\mathrm{VLAP}^{1}$    &      & $\surd$                                   & $\surd$                                   &          \textbf{56.7}/\textbf{78.8}/\textbf{86.2}     &        \textbf{50.7}/\textbf{72.1}/\textbf{82.8}        &     \textbf{61.0}/\textbf{70.0}      \\
$\mathrm{3D}$-$\mathrm{VLAP}^{2}$ & $\surd$   &      & $\surd$     & 45.3/70.5/79.7 & 50.3/74.4/89.0 & 51.5/62.7 \\
$\mathrm{3D}$-$\mathrm{VLAP}^{3}$   & $\surd$         & $\surd$        &         & 55.0/77.0/84.5 & 38.0/57.1/72.4 & 61.7/70.2 \\
$\mathrm{3D}$-$\mathrm{VLAP}$   & $\surd$      &        &        & 50.0/74.0/83.5 & 35.2/48.4/62.3 & 42.7/52.8 \\ \midrule
RA           &         &         &         & 0.6/3.9/7.2    & 4.1/11.2/28.3  & 0.0/11.3  \\ \bottomrule
\end{tabular}}
\end{table}

{
\bibliographystyle{IEEEtran}
\bibliography{references}
}
\newpage

\vfill

\end{document}